\documentclass[10pt,journal,onecolumn]{IEEEtran}
\IEEEoverridecommandlockouts
\usepackage{cite}
\usepackage{amsmath,amssymb,amsfonts}
\usepackage{algorithmic}
\usepackage{graphicx}
\usepackage{textcomp}
\usepackage{xcolor}
\usepackage{algorithm} 
\usepackage{multirow} 

\newtheorem{theorem}{Theorem}
\newtheorem{definition}{Definition}

\def\BibTeX{{\rm B\kern-.05em{\sc i\kern-.025em b}\kern-.08em
    T\kern-.1667em\lower.7ex\hbox{E}\kern-.125emX}}
\begin{document}

\title{Trustworthy Preference Completion in Social Choice
}

\author{Lei~Li,~\IEEEmembership{Senior~Member,~IEEE,}
	Minghe~Xue, Huanhuan~Chen,~\IEEEmembership{Senior~Member,~IEEE,}
    and~Xindong~Wu,~\IEEEmembership{Fellow,~IEEE,}
\IEEEcompsocitemizethanks{\IEEEcompsocthanksitem This work has been supported by the National Natural Science Foundation of China under grants 62076087 \& 91746209
and the Program for Changjiang Scholars and Innovative Research Team in University (PCSIRT) of the Ministry of Education of China under grant IRT17R32.
\IEEEcompsocthanksitem L. Li and M. Xue are with Key Laboratory of Knowledge Engineering with Big Data (Hefei University of Technology),
Intelligent Interconnected Systems Laboratory of Anhui Province (Hefei University of Technology),
and School of Computer Science and Information Engineering, Hefei University of Technology, Hefei, China.\protect\\
E-mail: lilei@hfut.edu.cn
\IEEEcompsocthanksitem H. Chen is with the School of Computer Science and Technology, University of Science and Technology of China.%
\IEEEcompsocthanksitem X. Wu is with Mininglamp Academy of Sciences, Mininglamp Technologies, Beijing, China,
Key Laboratory of Knowledge Engineering with Big Data (Hefei University of Technology),
and School of Computer Science and Information Engineering, Hefei University of Technology, Hefei, China.\protect\\}%
}

\maketitle

\begin{abstract}
As from time to time it is impractical to ask agents to provide linear orders over all alternatives,
 for these partial rankings it is necessary to conduct preference completion. Specifically,
the personalized preference of each agent over all the alternatives can be estimated
with partial rankings from neighboring agents over subsets of alternatives.
However, since the agents' rankings are nondeterministic,
where they may provide rankings with noise,
it is necessary and important to conduct the trustworthy preference completion.
Hence, in this paper firstly,
a trust-based anchor-kNN algorithm is proposed to find $k$-nearest trustworthy neighbors of the agent with
trust-oriented Kendall-Tau distances,
which will handle the cases when an agent exhibits irrational behaviors or provides only noisy rankings.
Then, for alternative pairs,
a bijection can be built from the ranking space to the preference space,
and its certainty and conflict can be evaluated based on a well-built statistical measurement Probability-Certainty Density Function.
Therefore,
a certain common voting rule for the first $k$ trustworthy neighboring agents based on certainty and conflict can be taken
to conduct the trustworthy preference completion.
The properties of the proposed certainty and conflict have been studied empirically,
and the proposed approach has been experimentally validated compared to state-of-arts approaches with several data sets.
\end{abstract}


\section{Introduction}

In a preference completion problem, with a set of agents (users) and a set of alternatives (items),
each agent (user) has his/her partial ranking over a subset of alternatives (items)
and the goal of this problem is to infer each agent (user)'s personalized ranking or preference over all the alternatives (items)
including those alternatives (items) the agent (user) has not yet handled.
Obviously, if the agents are required to rank all alternatives to form linear orders, there is no need to complete the preference at all.
However, from time to time it is impractical to ask agents to provide linear orders over all alternatives,
especially in big data environments \cite{JAR2020}.
For example, perhaps the agent doesn't know the status of some alternatives because there are too many alternatives,
which makes the agent hard to rank all of them.
Or perhaps some alternatives are uncomparable for a certain agent.
These all result in partial rankings, and it is necessary to introduce preference completion.

The preference completion problem has been applied to applications in many areas,
such as social choice, recommender system, etc \cite{Liu2011retrival}.
For example, in social choice, each voter (agent) can cast a ballot by a ranking over all candidates (alternatives),
or a partial ranking over some candidates (alternatives).
As for partial rankings, it is necessary to complete the preference,
and then these different rankings over all candidates can be aggregated by a certain voting rule to form a joint decision.
In a recommendation system, each user can rate some items by a partial ranking.
Then the task of the recommendation system is to predict the ranking for the user on the items
that have not been rated by him/her based on the existed rankings
and then the system can recommend top-$k$ items according to this prediction and partial rankings.
To satisfy this requirement, some existing technologies such as collaborative filtering approach,
neighborhood-based approaches are proposed to handle the preference completion.
In this paper, we focus on the neighborhood-based approaches, which perform better than others in the literature \cite{Liu2019near}.

Traditionally, in neighborhood-based preference completion,
it is first to find the near neighbors of each agent and then aggregate these neighbors' rankings
to produce the predicted preference \cite{Katz2018Non}.
Liu et al. \cite{Liu2008Eigenrank} initially propose a Kendall-Tau (KT) distance-based method to find the near neighbors of an agent,
where the Kendall-Tau distance is a measure of the similarity between two agents' rankings.
The shorter the Kendall-Tau distance, the more similar the two agents.
While in \cite{Liu2008Eigenrank}, they assume that agents' rankings are deterministic, which is unrealistic in many settings.
For example, an agent can exhibit irrational behaviors, or provide only noisy rankings.
Therefore, in the nondeterministic settings,
the method proposed in \cite{Liu2008Eigenrank} does hold reasonable results in finding the near neighbors.
To fix this problem, Liu et al. \cite{Liu2019near} propose an anchor-based algorithm to find neighbors of an agent under the presence of multiple levels of randomness.
With a salient feature that leverages the rankings of many other agents to determine the similarities of two agents,
it outperforms other widely used techniques including \cite{Liu2008Eigenrank}'s but inevitably losing the running efficiency
because the algorithm utilizes many other agents' ranking information to ignore the presence of randomness.
However, since the agents' rankings are nondeterministic, where they may provide their rankings under noisy environments,
it is natural to find a constant to measure the quality of rankings the agents made under uncertainty settings \cite{Donovan2005trust}.
More precisely, in finding neighbors of an agent, there is a strong desire to find the agents with rankings of high quality
and exclude the agents with rankings of low quality \cite{Knijnenburg2012Ins}.
Recent research has put forward a number of proposals for quantizing the quality of rankings \cite{Jamali2009Trustwalker}.
Hence, in this paper, we focus on utilizing this quality to improve the performance of preference completion on trustworthiness.

Technically, in a ranking pool gathered from agents,
these rankings between the alternative pair $A$ and $B$ can be aggregated to form the preference between $A$ and $B$.
Mathematically, a bijection can be built from the ranking space to the preference space for alternative pair $A$ and $B$.
Here, the ranking space consists of all the partial rankings on $A$ and $B$ from agents,
while the preference space consists of three-way preference between $A$ and $B$,
which includes preference (prefer $A$ to $B$, denoted as $P^+_{\textit{AB}}$),
dispreference (prefer $B$ to $A$, denoted as $P^-_{\textit{AB}}$),
and uncertainty (no preference between $A$ and $B$, denoted as $C^-_{\textit{AB}}$),
according to the trisecting and acting models of human cognitive behaviors \cite{JAR2020} \cite{journals/ijar/JingTaoYao20_2}.
Thus, the following three situations are distinguished:
\begin{itemize}
\item[-] The agents prefer alternative $A$ to alternative $B$,
which can be confirmed by high $P^+_{\textit{AB}}$, low $P^-_{\textit{AB}}$, and low $C^-_{\textit{AB}}$.

\item[-] The agents prefer alternative $B$ to alternative $A$,
which can be confirmed by low $P^+_{\textit{AB}}$, high $P^-_{\textit{AB}}$, and low $C^-_{\textit{AB}}$.

\item[-] The agents are uncertain about the preference between alternative pair $A$ and $B$,
i.e., $A$ and $B$ are unpreferred,
which can be confirmed by low $P^+_{\textit{AB}}$, low $P^-_{\textit{AB}}$, and high $C^-_{\textit{AB}}$.
\end{itemize}

It is obvious that when $C^-_{\textit{AB}}$ is low, the preference between $A$ and $B$ can be determined,
i.e., $A$ and $B$ are preferable.
Hence, the certainty of preference can be introduced to describe the trustworthiness of the preference,
which is denoted as $C^+_{\textit{AB}}$, and it can be calculated as $C^+_{\textit{AB}}=1-C^-_{\textit{AB}}$.
The certainty of preference can be taken as the subjective probability of the preference,
following the proposition that the certainty is the degree of belief that an individual has on the preference \cite{ATC2010}.
Hence, in this paper, the certainty based on a well-built statistical measurement can be evaluated,
which defines a bijection from ranking space to preference space,
enabling the aggregation of partial rankings via mapping them to
$($preference $P^+_{\textit{AB}},$ dispreference $P^-_{\textit{AB}},$ uncertainty $C^-_{\textit{AB}})$.
Our definition captures the following key properties:
\begin{enumerate}
\item[-] \textbf{Property 1}: Certainty $C^+_{\textit{AB}}$ increases
as the number of rankings between alternative pair $A$ and $B$ increases
for a fixed ratio of rankings from $A$ to $B$ and rankings from $B$ to $A$.

\item[-] \textbf{Property 2}: Certainty $C^+_{\textit{AB}}$ decreases
as the extent of conflict increases in the partial rankings between alternative pair $A$ and $B$.
\end{enumerate}

Our contributions in this paper can be summarized as follows:
\begin{itemize}
\item A trust-based anchor-kNN algorithm in finding $k$-nearest trustworthy neighbors of the agent with trust-oriented Kendall-Tau distances
is proposed to handle the preference completion problem trustworthily in the nondeterministic settings.

\item As pointed out in \cite{Hallinan2010},
it is necessary and important to introduce the certainty and conflict of the preference,
and from time to time the certainty and conflict of the preference are more important than the preference itself.
In this paper, a probability-based certainty and conflict are introduced under the above properties 1 \& 2,
to describe the trustworthiness of the preference,
and then the certainty and conflict can be used to conduct the trustworthiness preference completion.

\item We empirically study the properties of the proposed approach,
and experimentally validate the proposed approach compared to state-of-arts approaches with serval data sets.

\end{itemize}

This paper is organized as follows.
Section \ref{Background} reviews existing works on the Plackett-Luce model, Kendall-Tau distance, and anchor-kNN algorithm.
In Section \ref{Trust-based Anchor-kNN},
a trust-based anchor-kNN algorithm is proposed to find trustworthy neighboring agents in nondeterministic settings or noisy settings.
Then in Section \ref{Certainty and Preference Completion},
certainty and conflict of alternative pairs can be evaluated based on a well-built statistical measurement,
and a certain common voting rule can be taken for the trustworthy neighboring agents to conduct the preference completion with the certainty and conflict.
In addition, Section \ref{Empirical} studies empirically the properties of the proposed approach about certainty and conflict.
Moreover, Section \ref{Experiments} has been experimentally validated compared to state-of-arts approaches with serval data sets.
Finally, Section \ref{Conclusions} summarizes this paper and presents the future work.

\section{Background}\label{Background}

\subsection{Plackett-Luce Model}\label{Plackett-Luce model}

Given a set of $m$ alternatives and a set of $n$ agents, let $y({y_1},{y_2},...,{y_m})$ denote the latent features of the alternatives and $x({x_1},{x_2},...,{x_n})$ denote the latent features of the agents. The Plackett-Luce model\cite{Luce1959,Plackett1975} is a statistical model for ranking data. In the Plackett-Luce model, each alternative is assigned a positive value named utility. The greater this utility is, the more likely its corresponding alternative is ranked at a higher position \cite{Liu2019PL}. Hence, the Plackett-Luce model can be adopted to generate the rankings of agents. In \cite{Liu2019PL}, the realized utility for every alternative $j$ on agent $i$ is determined by $${u_{ij}}({x_i},{y_j}) = \theta ({x_i},{y_j}) + {\varepsilon _{i,j}},$$
$\theta ({x_i},{y_j})$ is agent $i$'s expected utility on alternative $j$ and can be determined by the closeness of the latent feature ${x_i}$ and ${y_j}$  measured by $$\theta ({x_i},{y_j}) = \exp ( - ||{x_i} - {y_j}|{|_2}).$$
${\varepsilon _{i,j}}$ is a zero mean independent random variable that follows a Gumbel distribution.
When the realized utilities set ${u_i}({u_i}_1,{u_{i2}},...{u_{im}})$ of the agent $i$ is obtained,
the agent $i$ ranks the alternatives in a decreasing order according to the realized utilities.
After repeating this for $n$ times, synthetic data sets of all the agents can be generated for experiments.
For more details, please refer to the following Algorithm \ref{alg1}.

\begin{algorithm}
	\caption{Sampling from Plackett-Luce}
\small{
	\label{alg1}
	\begin{algorithmic}[1]
		\REQUIRE a latent feature set on $n$ agents $x({x_1},{x_2},...,{x_n})$; a latent feature set $y({y_1},{y_2},...,{y_m})$ on $m$ alternatives.
		\ENSURE A data set of rankings $R\{ {R_1},{R_2}, \cdots {R_n}\} $
		\FOR{${x_i}$ in $x({x_1},{x_2},...,{x_n})$}
		\FOR{${y_j}$ in $y({y_1},{y_2},...,{y_m})$}
		\STATE Sample ${\varepsilon _{i,j}}$ follow a Gumbel distribution
		\STATE Compute ${u_{ij}}({x_i},{y_j}) = \exp ( - ||{x_i} - {y_j}|{|_2}) + {\varepsilon _{i,j}}$
		\ENDFOR
		\ENDFOR
        \FOR{${x_i}$ in $x({x_1},{x_2},...,{x_n})$}
		\FOR{$t = 1$ to $m$ }
		\STATE Choose an alternative ${y_{j}}$ from $y$ with a probability proportional to ${u_{{ij}}}$.
		\STATE $R_i \leftarrow R_i > {y_{j}}$ and $y \leftarrow y\backslash \{ {y_{j}}\} $
		\ENDFOR
        \ENDFOR
        \RETURN $R\{ {R_1},{R_2}, \cdots, {R_n}\} $
	\end{algorithmic}}
\end{algorithm}

\subsection{Kendall-Tau Distance}

Given two agents' rankings ${R_1}$ and ${R_2}$ over the same alternatives,
the Kendall-Tau distance can be introduced to measure the similarity of ${R_1}$ and ${R_2}$,
which is the total number of disagreements in pairwise comparisons between alternatives in linear orders.
For example, with the rank of the alternatives represented by $IR$,
if $j_1$ in $R_i$ is the top-ranked alternative, then $\emph{IR}_i(j_1)=1$.
The normalized Kendall-Tau distance between ${R_1}$ and ${R_2}$ is
\begin{eqnarray}
\!\!\!\!\emph{NK}({R_1},{R_2})=\frac{\sum\limits_{{j_1} \ne {j_2} \in {R_1}} {I\left(\prod_{k=1,2}({\emph{IR}_k}({j_1}) - {\emph{IR}_k}({j_2}))< 0\right)}}{{\left(\begin{array}{*{20}{c}}
		{|{R_1}|}\\
		2
		\end{array}\right)}}, \nonumber
\end{eqnarray}
where the $I(v)$ be an indicator that sets to $1$ if the argument $v$ is true; otherwise, it sets to $0$.

Moreover, if the rankings have not shared completely the same alternatives,
the intersection of the two alternative sets can be taken for computing the normalized Kendall-Tau distance.

\subsection{Anchor-kNN Algorithm}

Before the introduction of the anchor-kNN proposed in \cite{Liu2019PL}, we first present the idea of KT-kNN,
which simply uses the Kendall-Tau distance to find the agent's neighbors.
If the Kendall-Tau distance between two rankings $R_i$ and $R_j$ is small,
their latent feature of the agents $x_i$ and $x_j$ should be close,
i.e., the two agents have similar opinion on alternatives.
As this KT-kNN has not considered that the agents' preferences are nondeterministic and agents' rankings are made in noise setting,
different from KT-kNN, anchor-kNN uses other agents' (named as anchors) ranking data to determine the closeness of two agents rather than considering the two agents' ranking only.
The anchor-kNN develops a feature $F_{i,j}$ for agents $i$ and $j$ to represent the Kendall-Tau distance between $R_i$ and $R_j$,
i.e., ${F_{i,j}} = NK({R_i},{R_j})$. Then to measure the closeness of two agents denoted as
$D_{i,j}$, we use the sum of the difference between $F_{i,t}$ and $F_{j,t}$ to find the $k$-nearest neighbors, where $t$ is the third agent that belongs to all the other agents except agents $i$ and $j$. Please refer to Algorithm \ref{alg2} for more details.

\begin{algorithm} 
	\caption{Anchor-kNN} 
	\label{alg2} 
\small{
	\begin{algorithmic}[1] 
		\REQUIRE ${\{ R_j\} _{j \in n}}$, the number of the neighbors $k$, the latent feature $x_i$ on agent $i$ 
		\ENSURE $k$ neighbors near agent $i$ 
		\FOR{$i=0$ to $n-1$ }
		\FOR{$j=0$ to $n-1$ }
		\STATE 	Compute ${F_{i,j}} = NK({R_i},{R_j})$
		\ENDFOR
		\ENDFOR
        \FOR{$i=0$ to $n-1$ }
		\FOR{$j=0$ to $n-1$ }
        \STATE Compute $D({x_i},{x_j}) = \frac{1}{{n - 2}}\sum\limits_{t \ne i,j}^{} {|{F_{i,t}} - {F_{j,t}}|} $
        \ENDFOR
        \ENDFOR
        \STATE Find ${j_1}$,${j_2}$,$ \cdots $,${j_{n-1}}$ $\in [n]/\{ i\}$ such that
                $D({x_i},{x_{{j_1}}}) \le D({x_i},{x_{{j_2}}}) \cdots  \le D({x_i},{x_{j_{n - 1}}})$
        \RETURN ${X_{Anchor - kNN}} \leftarrow \{ {j_1},{j_2}, \cdots, {j_{\rm{k}}}\} $
	\end{algorithmic}}
\end{algorithm}

\section{Trust-based Anchor-kNN}\label{Trust-based Anchor-kNN}

Trust is the belief of one participant in another, based on their interactions \cite{Guo2014Merge,Guanfeng2010,Guanfeng2013},
with the extent to which a future action to be performed by the latter will lead to an expected outcome \cite{lei_roadmap14}.
In preference completion,
the trust can be used to measure the quality of the agent's rankings,
which can be obtained after evaluating trust ratings and/or other trust contextual information by trust management authorities \cite{lei_roadmap14}\cite{Lathia2008trustCF}\cite{Hwang2007TrustCF}.
Obviously, in nondeterministic settings or noise settings,
the trust value $trust(i)\in [0,1]$ can be adopted to discount the feature noted as $D_{i,j}$,
and then to find the $k$-nearest neighbors of the target agent ${x_i}$.
\begin{itemize}
\item[-] If $trust(i)\leq\epsilon_0$, the corresponding ranking is untrustworthy,
and it is justified that there is unnecessary to choose this agent $i$ as the neighbors of target agent,
i.e., this ranking can be eliminated from our data set, and this can improve the efficiency of algorithm,
where $\epsilon_0$ is the threshold to rule out the fuzziness of comparison.
\item[-] If $1-trust(i)\leq\epsilon_0$, the agent $i$ is a fairly trustworthy agent to be chosen.
\item[-] Without any prior additional information provided,
$trust(i)$ can be initially taken as $1$.
\end{itemize}

Based on the introduction of trust of the agents' rankings, the algorithm can be illustrated described in Algorithm \ref{alg3}
\footnote{In our experimental section, about synthetic data sets, without loss of generality,
$\{trust(i)\}$ are set according to the random variable ${\varepsilon _{i,j}}$ which follows a Gumbel distribution.}.

\begin{algorithm} 
	\caption{Trust-based Anchor-kNN} 
	\label{alg3} 
\small{
	\begin{algorithmic}[1] 
		\REQUIRE ${\{ R_j\} _{j \in n}}$, the trust set $\{trust(j) _{j \in n}\}$ on $n$ rankings, the number of the neighbors $k$, the latent feature $x_i$ on agent $i$
		\ENSURE $k$ neighbors near agent $i$ 
		\FOR{$i=0$ to $n-1$ }
		\FOR{$j=0$ to $n-1$ }
		\STATE Compute ${F_{i,j}} = NK({R_i},{R_j})$
		\ENDFOR
		\ENDFOR
        \FOR{$i=0$ to $n-1$ }
		\FOR{$j=0$ to $n-1$ }
        \STATE Compute $D({x_i},{x_j}) = \frac{{\frac{1}{{n - 2}}\sum\limits_{t \ne i,j} {|{F_{i,t}} - {F_{j,t}}|} }}{{trust(j)}}$
        \ENDFOR
        \ENDFOR
        \STATE Find ${j_1}$,${j_2}$,$ \cdots $,${j_{n-1}}$ $\in [n]/\{ i\}$ such that $D({x_i},{x_{{j_1}}}) \le D({x_i},{x_{{j_2}}}) \cdots  \le D({x_i},{x_{j_{n - 1}}})$
        \RETURN ${X_{Anchor - kNN}} \leftarrow \{ {j_1},{j_2}, \cdots, {j_{\rm{k}}}\} $
	\end{algorithmic}}
\end{algorithm}

\section{Certainty and Preference Completion}\label{Certainty and Preference Completion}

For all the rankings of arbitrary alternative pair $A$ and $B$ in the data set,
the certainty can be adopted to evaluate the degree of belief that the agent prefer $A$ to $B$ in alternative pair $A$ and $B$.
Technically, as the degree of belief that an agent has on the ranking \cite{lei_aaai10},
the certainty of ranking can be taken as the subjective probability of the ranking.
Following \cite{conf/esorics/Josang98},
a Probability-Certainty Density Function (PCDF) can be introduced to capture the subjective probability of the ranking.
However, unlike \cite{conf/esorics/Josang98},
following \cite{lei_aaai10} and \cite{journals/taas/WangS10}, in this paper
certainty is defined based on the PCDF to satisfy the mentioned properties 1 \& 2.

After obtaining the certainty, the preference completion with trustworthiness can be conducted.
Specifically, with the rankings of trustworthy neighbors obtained by trust-based anchor-kNN described in Algorithm \ref{alg3},
a certain common voting rule for the first $k$ trustworthy neighboring agents based on certainty can be taken
to conduct the preference completion.

Before the introduction of the formal definition on the PCDF-based certainty,
let's present some preliminary definitions firstly.

\subsection{Ranking Space}

The ranking space consists of all the weighted partial rankings on the alternative pair $A$ and $B$ from agents,
including
\begin{itemize}
\item[-] the rankings $\{O_{\textit{AB}}^{(i)}\}$ from $A$ to $B$ with weight $w_{\textit{AB}}^{(i)}$ for $O_{\textit{AB}}^{(i)}$,
where $n_{\textit{AB}}=\sum_iw_{\textit{AB}}^{(i)}$,
\item[-] the rankings $\{O_{\textit{BA}}^{(j)}\}$ from $B$ to $A$ with weight $w_{\textit{BA}}^{(j)}$ for $O_{\textit{BA}}^{(j)}$,
where $n_{\textit{BA}}=\sum_jw_{\textit{BA}}^{(j)}$,
and
\item[-] the unordered ones $\{O_{\overline{\textit{AB}}}^{(k)}\}$ between $A$ and $B$
with weight $w_{\overline{\textit{AB}}}^{(k)}$ for $O_{\overline{\textit{AB}}}^{(k)}$,
where $n_{\overline{\textit{AB}}}=\sum_kw_{\overline{\textit{AB}}}^{(k)}$,
$w_{\overline{\textit{AB}}}^{(k)}=w_{\overline{\textit{BA}}}^{(k)}$, and $O_{\overline{\textit{AB}}}^{(k)}=O_{\overline{\textit{BA}}}^{(k)}$.
\end{itemize}

\begin{definition}
Ranking space $$O=\{<n_{\textit{AB}},n_{\textit{BA}},n_{\overline{\textit{AB}}}>|n_{\textit{AB}}>0,
n_{\textit{BA}}>0,n_{\overline{\textit{AB}}}>0\}.$$
\end{definition}

\subsection{Preference Space}

Traditionally,
the uncertainty is usually ignored, and sometimes dispreference has been taken into account as well,
which leads to some disturbing results shown in empirical study section.
According to the trisecting and acting models of human cognitive behaviors \cite{conf/rskt/Yao09,journals/ijar/JingTaoYao20_2},
the preference space consists of three-way preference between alternatives,
which includes
\begin{itemize}
\item[-] preference $P^+_{\textit{AB}}$ (prefer $A$ to $B$),
\item[-] dispreference $P^-_{\textit{AB}}$ (prefer $B$ to $A$), and
\item[-] uncertainty $C^-_{\textit{AB}}$ (no preference between $A$ and $B$).
\end{itemize}

\begin{definition}
Preference space
\begin{eqnarray}
P&=&\{<P^+_{\textit{AB}},P^-_{\textit{AB}},C^-_{\textit{AB}}>|\\
&&P^+_{\textit{AB}}+P^-_{\textit{AB}}+C^-_{\textit{AB}}=1,\min\{P^+_{\textit{AB}},P^-_{\textit{AB}},C^-_{\textit{AB}}\}>0\}.\nonumber
\end{eqnarray}
\end{definition}

\subsection{Certainty of Rankings in Alternative Pairs}

The Bayesian inference \cite{Huanhuan1,Huanhuan2} here
is adopted to update the probability with the available information about the rankings in alternative pairs,
i.e. update the prior distribution to the posterior distribution \cite{Bayesian_Reliability,probability2003}.
Currently, the offline Bayesian inference has been utilized in this paper.
The Bayesian inference can also be applied to online/streaming scenario \cite{Huanhuan2014,Huanhuan2017}.

Let $x_{\textit{AB}}$, $x_{\textit{BA}}$ and $x_{\overline{\textit{AB}}}$
be the probability of rankings $\{O_{\textit{AB}}^{(i)}\}$, $\{O_{\textit{BA}}^{(j)}\}$ and $\{O_{\overline{\textit{AB}}}^{(k)}\}$, respectively,
where $x_{\overline{\textit{AB}}}=1-x_{\textit{AB}}-x_{\textit{BA}}$ and $X=<x_{\textit{AB}}, x_{\textit{BA}}>$. In addition, ${x_{AB}} \in [0,1]$, ${x_{BA}} \in [0,1]$ and ${x_{\overline {AB} }} \geq 0$, thus we then have ${x_{AB}} + {x_{BA}} \leq 1$.

Without any additional information,
the prior distribution $f(X|O)$ is a uniform distribution.
As the cumulative probability of a distribution within $[0,1]$ equals $1$,
the density of a PCDF has the mean value $1$ within $[0,1]$,
and this makes $f(X|O)=1$.

As the ranking sample $O$ conforms to a multinomial distribution \cite{probability2003,lei_aaai10},
we have
\begin{eqnarray}
f(O)=\frac{6(x_{\textit{AB}})^{n_{\textit{AB}}}(x_{\textit{BA}})^{n_{\textit{BA}}}(x_{\overline{\textit{AB}}})^{n_{\overline{\textit{AB}}}}}
{n_{\textit{AB}}!n_{\textit{BA}}!(n_{\overline{\textit{AB}}})!}.
\end{eqnarray}

As for posterior distribution $f(O|X)$,
it can be estimated as \cite{probability2003,lei_aaai10}
\begin{eqnarray}
&&f(O|X)=\frac{f(X|O)f(O)}{\int^1_0f(X|O)f(O)\mathrm{d}X}\\
&&=\frac{(x_{\textit{AB}})^{n_{\textit{AB}}}(x_{\textit{BA}})^{n_{\textit{BA}}}(x_{\overline{\textit{AB}}})^{n_{\overline{\textit{AB}}}}}
{\int^1_0(x_{\textit{AB}})^{n_{\textit{AB}}}(x_{\textit{BA}})^{n_{\textit{BA}}}
(x_{\overline{\textit{AB}}})^{n_{\overline{\textit{AB}}}}\mathrm{d}X}\nonumber
\end{eqnarray}

Then, the certainty can be determined by the deviations of posterior distribution from the prior distribution, i.e. uniform distribution.
Hence, we have the following definition about certainty.

\begin{definition}
The certainty of rankings $\{<n_{\textit{AB}},n_{\textit{BA}},n_{\overline{\textit{AB}}}>\}$
can be estimated as
\begin{eqnarray}
&&\!\!\!\!\!\!\!\!\!\!\!\!C^+_{\textit{AB}}=\frac{1}{2}\int_0^1|f(O|X)-f(X|O)|\mathrm{d}X\\
&&\!\!\!\!\!\!\!\!\!\!\!\!=\frac{1}{2}\int_0^1|\frac{(x_{\textit{AB}})^{n_{\textit{AB}}}
(x_{\textit{BA}})^{n_{\textit{BA}}}
(x_{\overline{\textit{AB}}})^{n_{\overline{\textit{AB}}}}}
{\int^1_0(x_{\textit{AB}})^{n_{\textit{AB}}}(x_{\textit{BA}})^{n_{\textit{BA}}}
(x_{\overline{\textit{AB}}})^{n_{\overline{\textit{AB}}}}\mathrm{d}X}-1|
\mathrm{d}X\nonumber,
\end{eqnarray}
where $\frac{1}{2}$ is to remove the double counting of the deviations.
\end{definition}

From this definition, we have $C^+_\textit{AB}=C^+_\textit{BA}$.

\subsection{Conflict of Rankings}

The conflict can be determined by the relatively difference between weighted rankings $n_\textit{AB}$ and $n_\textit{BA}$,
as in \cite{journals/taas/WangS10}. More specifically,
\begin{itemize}
\item[-] there is the largest conflict, when weighted rankings $n_\textit{AB}=n_\textit{BA}$;

\item[-] there is the smallest conflict, when weighted rankings $n_\textit{AB}=0$ or $n_\textit{BA}=0$;
\end{itemize}
Hence, we have the following definition about conflict.

\begin{definition}
The conflict of rankings $\{<n_{\textit{AB}},n_{\textit{BA}},n_{\overline{\textit{AB}}}>\}$
can be estimated as
\begin{eqnarray}
c_{\textit{AB}}=\min\{\frac{n_\textit{AB}}{n_\textit{AB}+n_\textit{BA}},\frac{n_\textit{BA}}{n_\textit{AB}+n_\textit{BA}}\}
\end{eqnarray}
\end{definition}

From this definition, we have $c_\textit{AB}=c_\textit{BA}$.

\subsection{Bijection from Ranking Space to Preference Space}

With Definitions 1, 2, 3 \& 4, the following definition can be introduced.
\begin{definition}
The bijection from ranking space $\{<n_{\textit{AB}},n_{\textit{BA}},n_{\overline{\textit{AB}}}>\}$
to preference space $\{<P^+_{\textit{AB}},P^-_{\textit{AB}},C^-_{\textit{AB}}>\}$
can be estimated as
\begin{eqnarray}
P^+_\textit{AB}&=&\frac{n_\textit{AB}}{n_\textit{AB}+n_\textit{BA}+n_{\overline{\textit{AB}}}}C^+_\textit{AB}\\
P^-_\textit{AB}&=&\frac{n_\textit{BA}}{n_\textit{AB}+n_\textit{BA}+n_{\overline{\textit{AB}}}}C^+_\textit{AB}\\
C^-_\textit{AB}&=&1-C^+_\textit{AB}
\end{eqnarray}
\end{definition}

\subsection{Trustworthy Preference Completion}

Before conducting the preference completion, let's firstly introduce a definition.

\begin{definition}
With preference space $\{<P^+_{\textit{AB}},P^-_{\textit{AB}},C^-_{\textit{AB}}>\}$,
the following conclusions can be obtained,
\begin{itemize}
\item[-] if $C^-_{\textit{AB}}\geq\epsilon_1$, alternatives $\textit{A}$ and $\textit{B}$ are unpreferred;
\item[-] if $C^-_{\textit{AB}}<\epsilon_1$,
\begin{itemize}
\item[-] if $P^+_{\textit{AB}}- P^-_{\textit{AB}}\geq\epsilon_2$, we prefer $\textit{A}$ to $\textit{B}$;
\item[-] if $P^-_{\textit{AB}}- P^+_{\textit{AB}}\geq\epsilon_2$, we prefer $\textit{B}$ to $\textit{A}$;
\item[-] otherwise, $\textit{A}$ and $\textit{B}$ are unpreferred;
\end{itemize}
\end{itemize}
where $\epsilon_1$ and $\epsilon_2$ are thresholds to rule out the fuzziness of comparison.
\end{definition}

In the existing work, with the rankings of neighbors obtained by anchor-kNN described in Algorithm \ref{alg2},
majority voting is adopted for the first $k$ neighbor agents to conduct the preference completion.

In this paper, firstly trust-based anchor-kNN described in Algorithm \ref{alg3} is introduced to locate the trustworthy neighboring agents.
Then, with the rankings of these trustworthy neighbors,
 ommon voting rules\footnote{common voting rules may include positional scoring rules, maximin, and Bucklin. For more details, please refer to \cite{JAR2020}.}, such as majority voting,
can be taken over the preferable alternative pair to determine the bias between the alternative pair
from the first $k$ neighbor agents (Please refer to Algorithm 4 below for more details).
Technically, for the alternative pair $\textit{A}$ and $\textit{B}$
with $C^-_{\textit{AB}}<\epsilon_1$ and $|P^+_{\textit{AB}}- P^-_{\textit{AB}}|\geq\epsilon_2$,
the bias between $P^+_\textit{AB}$ and $P^-_\textit{AB}$ in the first $k$ neighbor agents can be aggregated,
and then a new ranking with high certainty and low conflict can be obtained.
Hence, we can have the trustworthy preference completion.

\begin{algorithm}
	\caption{Certainty-based Preference Completion}
	\label{alg4}
\small{
	\begin{algorithmic}[1]
		\REQUIRE a latent feature set $y(y_1,y_2,...,y_m)$ on alternatives, neighbors rankings $R_1,R_2,...,R_k$.
		\ENSURE A complete ranking $R_i$ for target agent $x_i$
	    \STATE Compute $C^+_{j_1,j_2}$ for all ${j_1},{j_2} \in m$
		\STATE Compute $P_{j_1,j_2}$ for all ${j_1},{j_2} \in m$
		\STATE /*aggregate the $k$ neighbors rankings to get the rating matrix $v_{j_1,j_2}\leftarrow 0$ for every pair of alternatives*/
		\FOR{$all j_1,j_2 \in m$ and $R_i in \{R_1,R_2,...R_k\}$ }
        \STATE ${v_{{j_1},{j_2}}} \leftarrow {v_{{j_1},{j_2}}} \pm  1$, if $y_{j_1}$ and $y_{j_1}$  in  $R_i$, $y_{j_1}$ rank before or rank after $y_{j_1}$
		\ENDFOR
		\STATE /*aggregate the Rating Matrix for every alternatives, $v_{{j_i}} \leftarrow 0$*/
        \FOR{$all j_i \in m$ and $all j_t \in m$ }
        \STATE ${v_{{j_i}}} = {v_{{j_i}}} + {v_{{j_i},{j_t}}}*{P_{{j_i},{j_t}}}$, if $C^-_{\textit{AB}}<\epsilon_1$ and $|P^+_{\textit{AB}}- P^-_{\textit{AB}}|\geq\epsilon_2$
		\ENDFOR
        \STATE Rank $\{ v_{2{{j_1}}},v_{2{{j_2}}}, \cdots ,v_{2{{j_m}}}\} $
        \STATE Get the complete rank ${R_i} \leftarrow \{ {y_{{i_1}}},{y_{{i_2}}}, \cdots ,{y_{{i_m}}}\}  $
        \RETURN $R_i$
	\end{algorithmic}}
\end{algorithm}

\section{Empirical Studies on Properties of Certainty}\label{Empirical}


\subsection{Increasing Rankings with Fixed Conflict}

\begin{figure}[htbp]
\centerline{\includegraphics[scale=0.55]{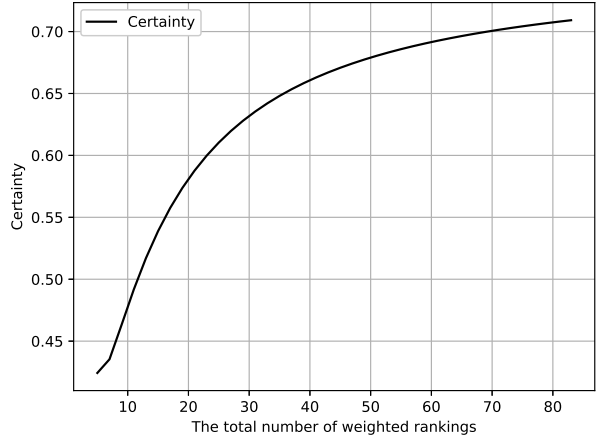}}
\caption{Certainty increases with $n_\textit{AB}+n_\textit{BA}$ when $\frac{n_\textit{AB}}{n_\textit{AB}+n_\textit{BA}}$ and $n_{\overline{\textit{AB}}}$ is fixed.}
\label{Figure1}
\end{figure}

Fig. \ref{Figure1} plots how certainty $C^+_\textit{AB}$ varies with weighted rankings $n_\textit{AB}$ and $n_{\overline{\textit{AB}}}$
under fixed conflict $c_\textit{AB}$.

This should confirm the Property 1.

\begin{theorem}
As for fixed $\frac{n_\textit{AB}}{n_\textit{AB}+n_\textit{BA}}$ and $n_{\overline{\textit{AB}}}$,
the certainty $C^+_\textit{AB}$ increases with $n_\textit{AB}+n_\textit{BA}$.\label{theorem1}
\end{theorem}

\noindent {\textit{Proof}:}
Let $\frac{n_\textit{AB}}{n_\textit{AB}+n_\textit{BA}}=\alpha$, $n_\textit{AB}+n_\textit{BA}=\beta$,
and
\begin{eqnarray}
f(\bullet)=\frac{(x_{\textit{AB}})^{n_{\textit{AB}}}
(x_{\textit{BA}})^{n_{\textit{BA}}}
(1-x_{\textit{AB}}-x_{\textit{BA}})^{n_{\overline{\textit{AB}}}}}
{\int^1_0(x_{\textit{AB}})^{n_{\textit{AB}}}(x_{\textit{BA}})^{n_{\textit{BA}}}
(1-x_{\textit{AB}}-x_{\textit{BA}})^{n_{\overline{\textit{AB}}}}\mathrm{d}X}.\nonumber
\end{eqnarray}
Then we have
\begin{eqnarray}
C^+_\textit{AB}=\frac{1}{2}\int_0^1|f(\bullet)-1|\mathrm{d}X.
\end{eqnarray}

As in \cite{journals/taas/WangS10}, $x_1$, $x_2$, $x_3$, $x_4$ can be defined, such that
$f(x_1)=f(x_2)=f(x_3)=f(x_4)=1$ and
\begin{eqnarray}
C^+_\textit{AB}=\int_{x_1}^{x_2}\int_{x_3}^{x_4}\left[f(\bullet)-1\right]
\mathrm{d}x_{\textit{AB}}\mathrm{d}x_{\textit{BA}},
\end{eqnarray}
where $x_1$, $x_2$, $x_3$, $x_4$ are functions of $\beta$. Then
\begin{eqnarray}
\frac{\partial C^+_\textit{AB}}{\partial \beta}&=&
\frac{\partial x_2}{\partial \beta}\int_{x_3}^{x_4}\left[f(x_2)-1\right]
\mathrm{d}x_{\textit{AB}}\nonumber\\
&&-\frac{\partial x_1}{\partial \beta}\int_{x_3}^{x_4}\left[f(x_1)-1\right]
\mathrm{d}x_{\textit{AB}}\nonumber\\
&&+\int_{x_1}^{x_2}\frac{\partial }{\partial \beta}\int_{x_3}^{x_4}\left[f(\bullet)-1\right]
\mathrm{d}x_{\textit{AB}}\mathrm{d}x_{\textit{BA}}\\
&=&\int_{x_1}^{x_2}\frac{\partial }{\partial \beta}\int_{x_3}^{x_4}\left[f(\bullet)-1\right]
\mathrm{d}x_{\textit{AB}}\mathrm{d}x_{\textit{BA}}\label{eqn13}
\end{eqnarray}
where
\begin{eqnarray}
&&\frac{\partial }{\partial \beta}\int_{x_3}^{x_4}\left[f(\bullet)-1\right]
\mathrm{d}x_{\textit{AB}}\\
&=&\frac{\partial x_4}{\partial \beta}\left[f(x_4)-1\right]-\frac{\partial x_3}{\partial \beta}\left[f(x_3)-1\right]\nonumber\\
&&
+\int_{x_3}^{x_4}\frac{\partial }{\partial \beta}\left[f(\bullet)-1\right]\mathrm{d}x_{\textit{AB}}\nonumber\\
&=&\int_{x_3}^{x_4}\frac{\partial }{\partial \beta}\left[f(\bullet)-1\right]\mathrm{d}x_{\textit{AB}}
\end{eqnarray}
Following Lemma 9 in \cite{journals/taas/WangS10},
we have
\begin{eqnarray}
\frac{\partial }{\partial \beta}\int_{x_3}^{x_4}\left[f(\bullet)-1\right]
\mathrm{d}x_{\textit{AB}}>0
\end{eqnarray}

With Eq. (\ref{eqn13}), we have
\begin{eqnarray}
\frac{\partial C^+_\textit{AB}}{\partial \beta}>0
\end{eqnarray}
This confirms the results of Theorem \ref{theorem1}.

\subsection{Increasing Conflict with Fixed Rankings}

\begin{figure}[htbp]
\centerline{\includegraphics[scale=0.55]{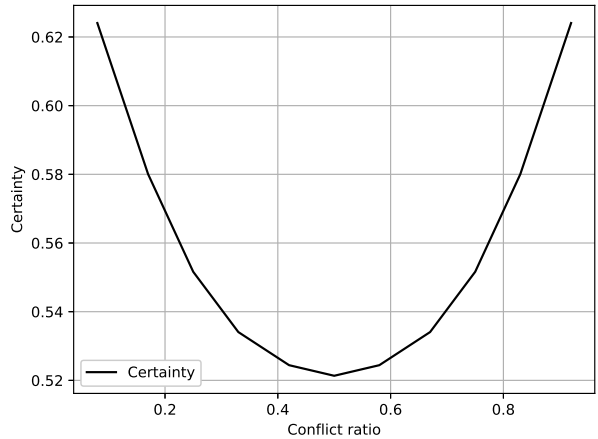}}
\caption{\emph{Certainty is concave when ${n_{AB}} + {n_{BA}} + {n_{\overline {AB} }}$ and $n_{\overline{{AB}}}$ is fixed, minimum occurs at $n_{AB}=n_{BA}$}}
\label{Figure2}
\end{figure}

Fig. \ref{Figure2} plots how certainty $C^+_\textit{AB}$ varies with weighted rankings $n_\textit{AB}$ and $n_{\overline{\textit{AB}}}$
under the fixed summation of $n_{\textit{AB}}+n_{\textit{BA}}$ and the fixed $n_{\overline{\textit{AB}}}$.
This should confirm the Property 2.

\begin{theorem}
As for fixed $n_{\overline{\textit{AB}}}$,
the certainty $C^+_\textit{AB}$ is decreasing with $n_\textit{AB}\leq n_\textit{BA}$,
and increasing with $n_\textit{AB}\geq n_\textit{BA}$.
\end{theorem}

\noindent {\textit{Proof}:}
The details of validation process can be omitted here,
as it is similar to one in the proof of Theorem \ref{theorem1}.
More specifically,
with removing the absolute sign and then differentiating it,
it can be proved that the derivation is negative for $n_\textit{AB}\leq n_\textit{BA}$,
and positive for $n_\textit{AB}\geq n_\textit{BA}$.

\section{Experiments}\label{Experiments}

This section reports the experimental results on synthetic data set and a real-world data set,
with the anchor-based kNN algorithm proposed in \cite{Liu2019near} and our trust-based anchor kNN algorithm.
After getting the $k$-nearest neighbors, then the trustworthy preference completion algorithm can be implemented.
Through the experiments for finding $k$-nearest neighbors and preference completion on the two algorithms
using not only the synthetic data set but also the real-world data set,
the superiority of the proposed method over non-trust anchor-based kNN methods can be demonstrated.
Further, the experiments also verify the effectiveness of certainty-based preference completion algorithm.

\subsection{Data Sets}

As discussed above, the experiments use two forms of data set to evaluate two algorithms' performance.
\begin{itemize}
\item[-] One type of data set is the synthetic data set created by the sampler using a Plackett-Luce model.
\item[-] The other type of data set is the Irish Election Data Set obtained from Preflib.
The produced synthetic data set has over 20000 numbers of agent rankings on the set of items with 20 numbers.
Each ranking has its own trust according to the random variable which followed a Gumbel distribution.
The Irish Election Data Set obtained from Preflib contains an uncomplete record of votes for two separate elections held in Dublin,
Ireland in 2002, which contains the Dublin North, West, and Meath data sets.
In the experiments the subset of the West data set which contains 29,988 over 9 alternatives is adopted.
Because of the memory capacity, 8700 agents over 9 alternatives are adopted.
\end{itemize}
\subsection{Evaluation Metric}

\begin{itemize}
\item[-] \textbf{Evaluation Metric 1}:
A widely used approach to measure the error between observed value and the ground Truth value is the Root Mean Squared Error (RMSE).
In the experiments, RMSE is defined as:

\begin{equation}
\emph{RMSE} = \sqrt {\frac{1}{k}{{({x_i} - {x_0})}^2}}
\end{equation}
where $k$ is the number of neighbors which are required to get,
${x_i}$ is the latent feature of the neighbor, and
${x_0}$ is the latent feature of the target agent.

\item[-] \textbf{Evaluation Metric 2}: In addition, a bias is adopted to measure the degree of bias of the predicted ranking.
\begin{equation}
B(\emph{rank}) = \frac{{\sum\limits_{i=0}^{i = m - 2} {\sum\limits_{j = i + 1}^{j = m - 1} {{P^+_{\emph{index}[R[i]],\emph{index}[R[j]]}}} } }}{{\sum\limits_{0}^{\frac{1}{2}(len(\emph{rank})*len(\emph{rank}))} {\max {{\{ {P^+_{i,j}}\} }_{i,j \in m}}} }}
\end{equation}

\item[-] \textbf{Evaluation Metric 3}: Then the precision can be evaluated with a bias and a Kendall-Tau distance
considered to measure the distance between the predicted rank and the original rank as well as the bias of the predicted rank
\begin{equation}
Pre = \emph{weight}_A*\emph{precision@5} + \emph{weight}_B*B(\emph{rank})
\end{equation}
\begin{equation}
\emph{precision@5} = 1 - \frac{{KT}}{{N({R_p} \cap {R_0})*(N({R_p} \cap {R_0}) - 1)}}
\end{equation}
where the ${N({R_p} \cap {R_0})}$ is the number of items in  $ {R_p} \cap {R_0}$.
\end{itemize}

\subsection{Experimental Results}

In this subsection, according to the evaluation measures separately,
first we present the results for the synthetic data set we create, and then the results for the Irish Election Data Set.

\begin{figure}[htbp]
\centerline{\includegraphics[scale=0.55]{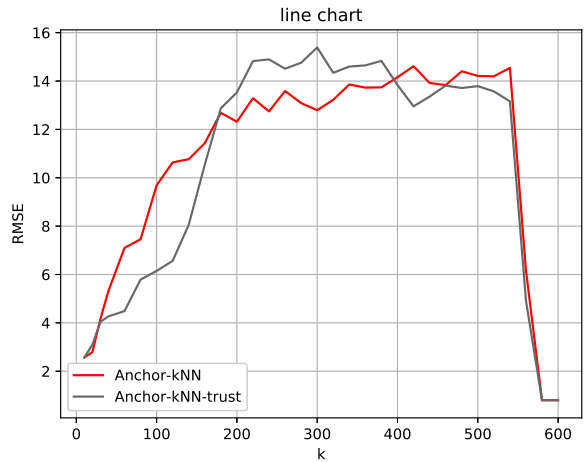}}
\caption{Comparison of RMSE for trust-based anchor-kNN and anchor-kNN.}
\label{fig3}
\end{figure}

\emph{Synthetic data set:} As shown in Fig. \ref{fig3},
when the number of neighbor $k$ is less than 180, the proposed trust-based anchor-kNN has a lower RMSE (evaluation metric 1) than the anchor-kNN,
while when $k$ is in the range [180,400], the proposed algorithm does not perform better than the anchor-kNN proposed in \cite{Liu2019near}.
That's because when there are more and more neighbors to be considered,
the proposed algorithm prefers to choose those neighbors whose trust are as large as possible
and the distance with the finding agent should be as small as possible rather than only consider the distance as the anchor-kNN do.
Hence, in some case the proposed algorithm does not perform better under the evaluation measure RMSE
which only considers the latent distance between the neighbors and the finding agent.
To sum up, the proposed algorithm can reduce the error when the neighbor's number is not too large.

\begin{figure}[htbp]
\centerline{\includegraphics[scale=0.55]{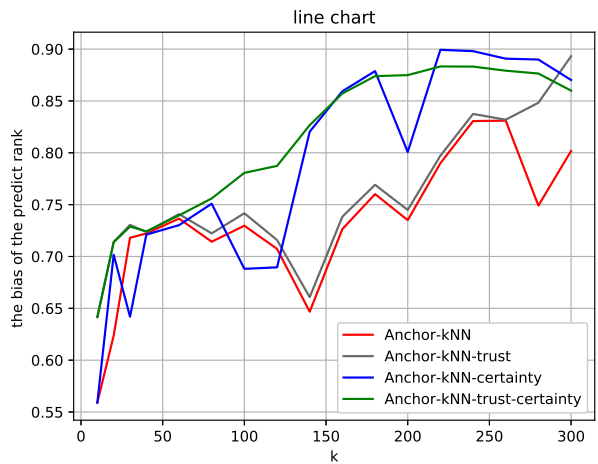}}
\caption{Comparison of the bias of the predict rank for Anchor-kNN, Anchor-kNN with trust, Anchor-kNN with certainty, Anchor-kNN with trust and certainty.}
\label{fig4}
\end{figure}

\begin{figure}[htbp]
\centerline{\includegraphics[scale=0.55]{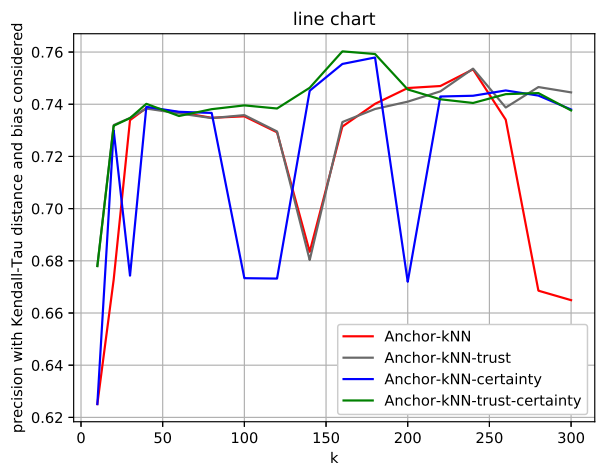}}
\caption{Comparison of the precision with Kendall-Tau distance and bias considered of the predict rank for Anchor-kNN, Anchor-kNN with trust, Anchor-kNN with certainty, Anchor-kNN with trust and certainty.}
\label{fig5}
\end{figure}

With the $k$ neighbors through the trust-based anchor-kNN or the anchor-kNN,
the new complete ranking can be predicted according to the neighbors' rankings.
As Section 4 mentioned, the certainty can be used during the process of preference completion.
Figs. \ref{fig4} and \ref{fig5} show the result for preference completion
following the step for finding the $k$ nearest neighbors in Section 3.

Fig. \ref{fig4} shows the charts comparing the bias (evaluation metric 2) of the predicted ranking after preference completion using different method.
\begin{itemize}
\item[-] The red line is the original method using anchor-kNN to find the $k$ neighbors and using the traditional algorithm without certainty considered to finish the preference completion.
\item[-] The blue line uses the Algorithm 4 with certainty considered to finish the preference completion.
\item[-] The gray line uses trust-based anchor kNN to find the $k$ neighbors as Algorithm 3 described.
\item[-] The green line not only uses the trust-based anchor kNN, but also uses the preference completion with certainty considered.
\end{itemize}
Fig. \ref{fig4} shows that Algorithms 3 and 4 can both significantly outperform the original method according to the bias of the predicted ranking.

Fig. \ref{fig5} shows the charts comparing the precision (evaluation metric 3) with bias and Kendall-Tau distance considered.
Similar with Fig. \ref{fig4}, the lines with different color represent the different methods.
We can see that the method using Algorithms 3 and 4 together
also clearly outperforms all other method including the method using Algorithm 3 or Algorithm 4
only and the original methods not using Algorithms 3 and 4.
It also should be noted that the red line, the blue line and the gray line significantly decrease in some $k$ value.
This because some distrusted agent's ranking which has low bias was to be put in the neighbors' team by the anchor-kNN.
While the proposed method using Algorithms 3 and 4 moves smoothly
since the proposed method reject those rankings which were not trusted and have low bias.
It can come to a conclusion that the proposed algorithm performs better than the other method and can address exceptional situation.


\emph{Real data set: }For the real data set -- Irish Election Data Set, the trust for every agent's ranking is not known,
so we only compare the performance under the evaluation metrics 2 and 3 between the traditional preference completion algorithm and the preference completion algorithm with certainty considered.

\begin{figure}[htbp]
\centerline{\includegraphics[scale=0.55]{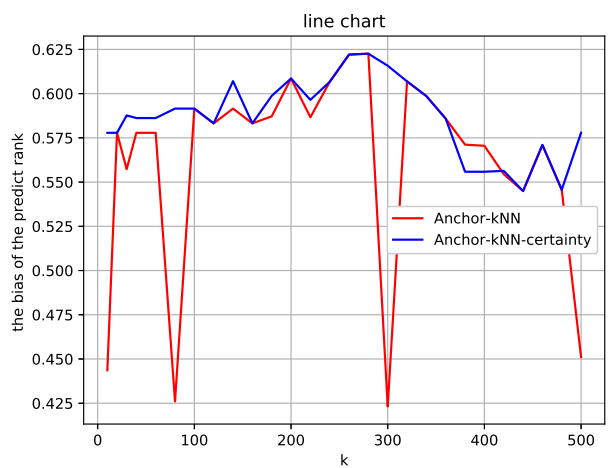}}
\caption{Comparison of the bias of the predict rank for Anchor-kNN and Anchor-kNN with certainty.}
\label{fig6}
\end{figure}

\begin{itemize}
\item[-] As shown in Fig. \ref{fig6}, the red line represents the bias (evaluation metric 2) of the predicted ranking
after using anchor-kNN for finding $k$ neighbors and using traditional preference completion algorithm
for produce a predicted completion ranking.
\item[-] While the blue one represents the bias of the predicted ranking
after using the same anchor-kNN and using the preference completion algorithm with certainty considered as Algorithm 4 described.
It is very clear that the proposed method using Algorithm 4 performs better than the traditional method.
\item[-] In addition, it also noted that the red one is very twisting and the bias at some point is very low.
This is because some agent's ranking which has low bias still be put into the neighbors' team.
And the proposed algorithm can avoid this situation.
\end{itemize}

\begin{figure}[htbp]
\centerline{\includegraphics[scale=0.55]{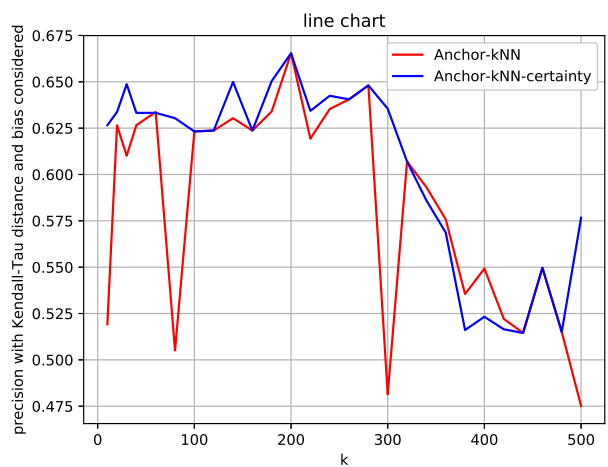}}
\caption{Comparison of the precision with Kendall-Tau distance and bias considered of the predict rank for Anchor-kNN and Anchor-kNN with certainty.}
\label{fig7}
\end{figure}

Fig. \ref{fig7} shows the comparison of the precision (evaluation metric 3) between the traditional preference completion algorithm
and the preference completion algorithm with certainty considered.
Similar with Fig. \ref{fig6}, the traditional one is very twisting and has steep decline and rise.
While the proposed algorithm not has such twisting bias and has the larger value of the precision.


\section{Conclusion and Future Work}\label{Conclusions}

Due to the fact that the agents' rankings are nondeterministic,
where they may provide their rankings under noisy environments,
it is necessary and important to conduct trustworthiness preference completion.
Hence, in this paper firstly, with trust-oriented Kendall-Tau distances,
a trust-based anchor-kNN algorithm is proposed to find trustworthy neighboring agents in nondeterministic settings or noise settings.
Then, for alternative pairs
a bijection can be built from the ranking space to the preference space,
and its certainty and conflict can be evaluated based on a well-built statistical measurement Probability-Certainty Density Function.
Therefore,
a certain common voting rule, such as majority voting, can be taken for the first $k$ trustworthy neighboring agents based on certainty and conflict
to conduct the preference completion.
More specifically, the ranking with high certainty and low conflict can be obtained during preference completion.
The properties of the proposed approach about certainty and conflict have been studied empirically,
and the proposed approach has been experimentally validated compared to state-of-arts approaches with serval data sets.

As in real applications,
the data are usually unbalanced \cite{Huanhuan2016},
i.e. some alternative pairs have a lot of rankings,
while some alternative pairs only have few rankings.
In our future work,
we will propose algorithms to handle unbalance preference completion both effectively and efficiently.

\begin{IEEEbiography}[{\includegraphics[width=1.1in,height=1.85in,clip,keepaspectratio]{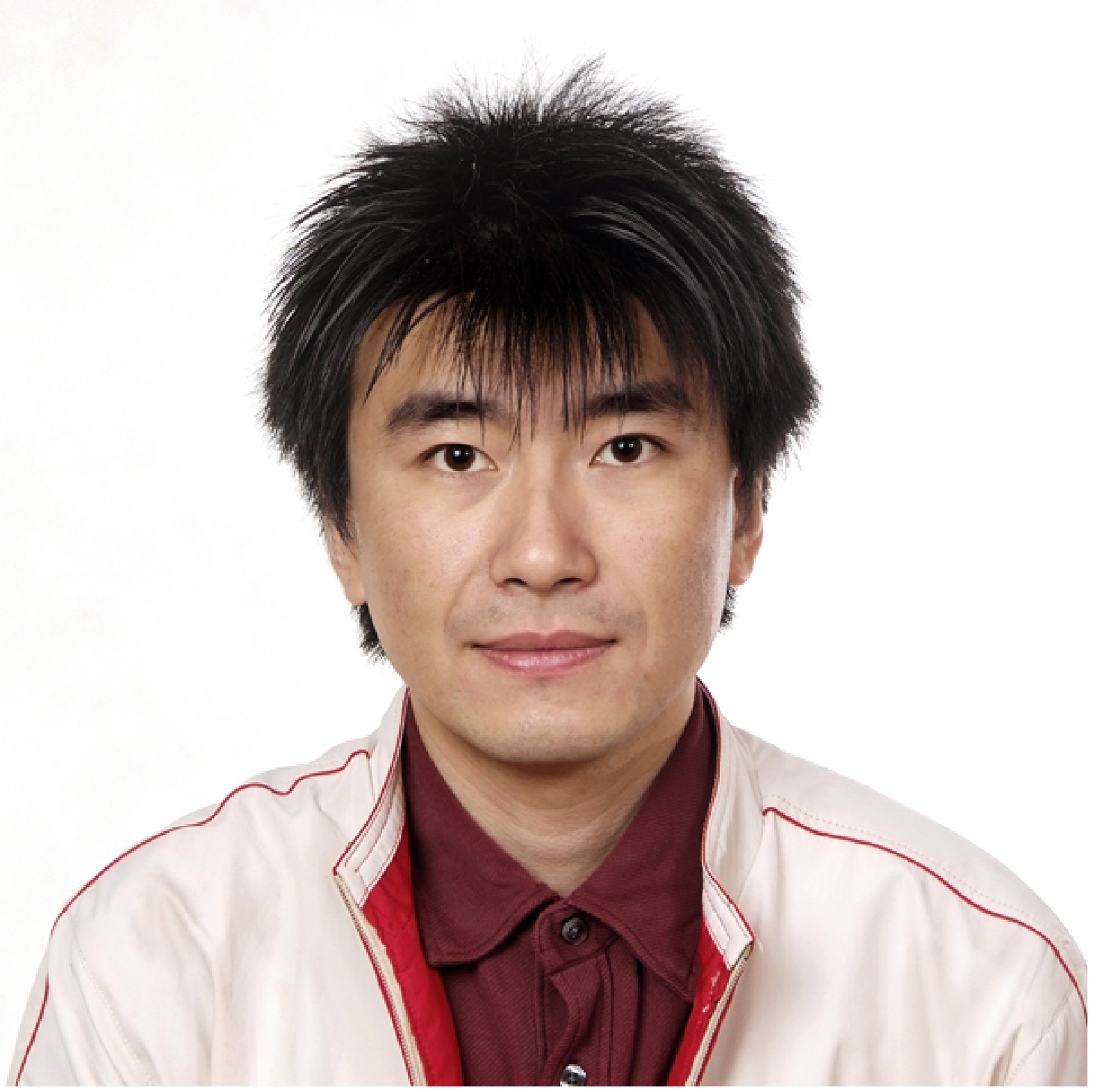}}]{Lei Li}
is currently an Associate Professor of computer science and technology at Key Laboratory of Knowledge Engineering with Big Data (Hefei University of Technology), Intelligent Interconnected Systems Laboratory of Anhui Province (Hefei University of Technology), and School of Computer Science and Information Engineering, Hefei University of Technology, Hefei, China. His research interests include data mining, social computing and graph computing. He received his Bachelor's degree in information and computational science from Jilin University, Changchun, China, in 2004, his Master's degree in applied mathematics from the Memorial University of Newfoundland, St. John's, Canada, in 2006, and his Ph.D. degree in computing from Macquarie University, Sydney, Australia, in 2012. He is a Senior Member of the IEEE.
\end{IEEEbiography}

\begin{IEEEbiography}[{\includegraphics[width=1.1in,height=1.85in,clip,keepaspectratio]{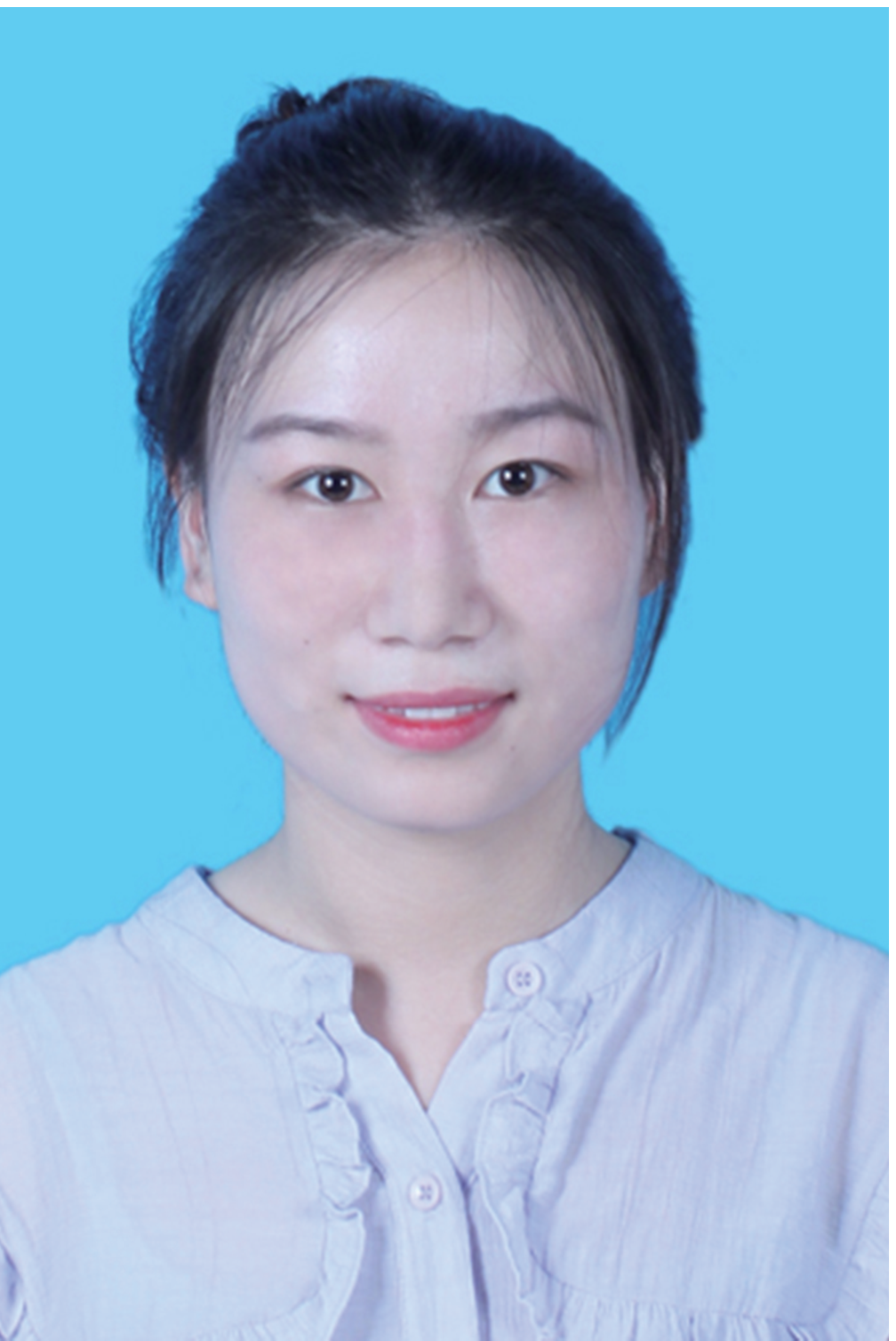}}]{Minghe Xue}
is currently working toward the M.S. degree at the School of Computer Science
and Technology, Hefei University of Technology, China. She received the B.S. degree from Hefei University of Technology, Anhui, China, in 2019. Her current research interests include data mining and intelligent computing.
\end{IEEEbiography}

\begin{IEEEbiography}[{\includegraphics[width=1.1in,height=1.85in,clip,keepaspectratio]{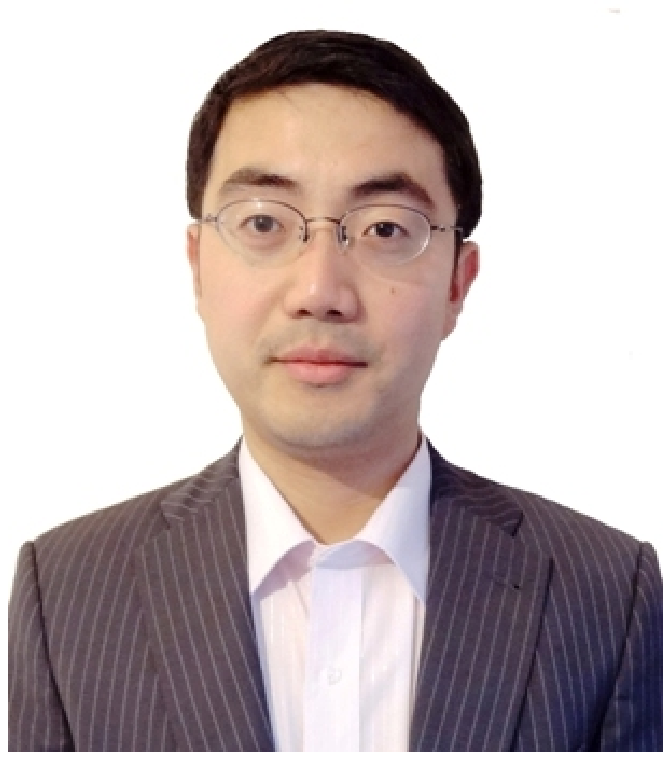}}]{Huanhuan Chen} (Senior Member, IEEE) received the B.Sc. degree from the University of Science and Technology of China (USTC), Hefei, China, in 2004, and the Ph.D. degree in computer science from the University of Birmingham, Birmingham, U.K., in 2008.

He is currently a Full Professor with the School of Computer Science and Technology, USTC. His current research interests include neural networks, Bayesian inference, and evolutionary computation. He was the recipient of the 2015 International Neural Network Society Young Investigator Award, the 2012 IEEE Computational Intelligence Society Outstanding Ph.D. Dissertation Award, the IEEE TRANSACTIONS ON NEURAL NETWORKS Outstanding Paper Award (bestowed in 2011 and only one paper in 2009), and the 2009 British Computer Society Distinguished Dissertations Award. He is currently an Associate Editor for the IEEE TRANSACTIONS ON NEURAL NETWORKS AND LEARNING SYSTEM, and the IEEE TRANSITION ON EMERGING TOPICS IN COMPUTATIONAL INTELLIGENCE.
\end{IEEEbiography}

\begin{IEEEbiography}[{\includegraphics[width=1.1in,height=1.85in,clip,keepaspectratio]{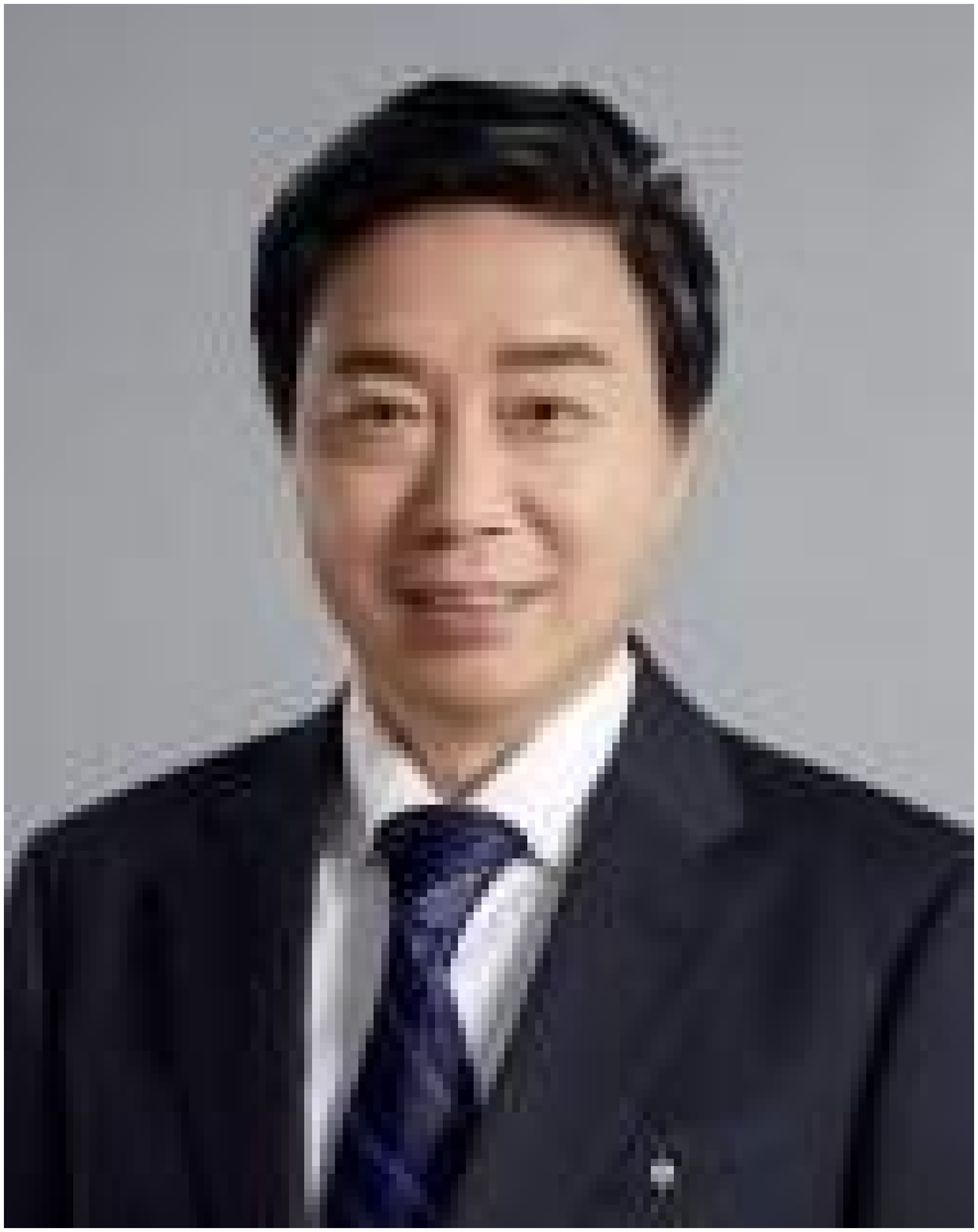}}]{Xindong Wu} is the chief scientist in Mininglamp Academy of Sciences, Mininglamp Technologies, Beijing, China, and a Chang Jiang Scholar in the School of Computer Science and Information Engineering at the Hefei University of Technology, China. His research interests include data mining, knowledge-based systems, and web information exploration. He received his Ph.D. in artificial intelligence from the University of Edinburgh. He is the Steering Committee Chair of the IEEE International Conference on Data Mining (ICDM), the Editor-in-Chief of Knowledge and Information Systems, and Editor-in-Chief of the Springer book series, Advanced Information and Knowledge Processing (AIKP). He is Fellow of IEEE and the AAAS.
\end{IEEEbiography}

\end{document}